\documentclass[letterpaper, 10 pt, conference]{ieeeconf}  % Comment this line out if you need a4paper

\IEEEoverridecommandlockouts                              % This command is only needed if 
\usepackage{graphics} % for pdf, bitmapped graphics files
\usepackage{epsfig} % for postscript graphics files
\usepackage{mathptmx} % assumes new font selection scheme installed
\usepackage{times} % assumes new font selection scheme installed
\usepackage{amsmath} % assumes amsmath package installed
\usepackage{amssymb}  % assumes amsmath package installed
\usepackage{algorithm}
\usepackage{color}
\usepackage{xcolor}
\usepackage{algpseudocode}
\usepackage{tabularx}
\usepackage{multirow}
\usepackage{booktabs}
\usepackage{caption}
\usepackage{pifont}
\newcommand{\xmark}{\ding{55}}%
\newcommand{\cmark}{\ding{51}}%
\newcommand\norm[1]{\left\lVert#1\right\rVert}

\title{\LARGE \bf
Local Non-Cooperative Games with Principled Player Selection \\ for Scalable Motion Planning}

\author{Makram Chahine$^{1}$, Roya Firoozi$^{2}$, Wei Xiao$^{1}$, Mac Schwager$^{2}$ and Daniela Rus$^{1}$% <-this % stops a space
\thanks{$^{1}$ Computer Science and Artificial Intelligence Lab, Massachusetts Institute of Technology \texttt{{\small \{chahine, weixy, rus\}@mit.edu}}}
\thanks{$^{2}$ Department of Aeronautics \& Astronautics, Stanford University \texttt{{\small \{rfiroozi, schwager\}@stanford.edu}}}
\thanks{This work was supported in part by ONR grant N00014-18-1-2830. Toyota Research Institute provided funds to support this work.  The second author was supported on an ASEE eFellows fellowship.}
}
\begin{document}

\maketitle
\thispagestyle{empty}
\pagestyle{empty}

%%%%%%%%%%%%%%%%%%%%%%%%%%%%%%%%%%%%%%%%%%%%%%%%%%%%%%%%%%%%%%%%%%%%%%%%%%%%%%%%
\begin{abstract}
Game-theoretic motion planners are a powerful tool for the control of interactive multi-agent robot systems. Indeed, contrary to predict-then-plan paradigms, game-theoretic planners do not ignore the interactive nature of the problem, and simultaneously predict the behaviour of other agents while considering change in one's policy. This, however, comes at the expense of computational complexity, especially as the number of agents considered grows. In fact, planning with more than a handful of agents can quickly become intractable, disqualifying game-theoretic planners as possible candidates for large scale planning. In this paper, we propose a planning algorithm enabling the use of game-theoretic planners in robot systems with a large number of agents. Our planner is based on the reality of locality of information and thus deploys local games with a selected subset of agents in a receding horizon fashion to plan collision avoiding trajectories. We propose five different principled schemes for selecting game participants and compare their collision avoidance performance. We observe that the use of Control Barrier Functions for priority ranking is a potent solution to the player selection problem for motion planning.
\end{abstract}

%%%%%%%%%%%%%%%%%%%%%%%%%%%%%%%%%%%%%%%%%%%%%%%%%%%%%%%%%%%%%%%%%%%%%%%%%%%%%%%%
\section{Introduction}

While interactions among multiple mobile agents can be thoroughly modeled using game theory, the complexity of game-theoretic planners grows cubically with the number of the agents considered \cite{leCleach2022algames}. Thus, in large scale settings where a large number of interactive agents are present, game-theoretic planning is intractable \cite{Daskalakis2009}. The typical procedure in large scale settings is to identify the closest neighbors and perform local game-theoretic planning while ignoring the rest of the agents in the scene \cite{Schwarting2019}. This neighbor selection scheme is justified in structured environments such as driving, where the geometry of the road and driving rules ensure agents closest are generally the most important to consider. However, in congested settings, taking the spatially closest neighbors into account  (e.g. all agents in a given radius) might still result in a computationally expensive planning scheme. Furthermore, in unstructured environments such as quadrotor flight, the validity of the argument for planning with the closest agents is highly questionable.

In this work we study the problem of game-theoretic motion planning for systems with a number of agents beyond the computational feasibility of a full game solution. The investigation aims to provide insights into this question on two levels. First, we aim to evaluate the capacity of receding horizon local game-theoretic planners to achieve satisfactory trajectories in crowded environments, when the maximum number of agents considered for planning is constrained. Trajectories in this setting are obtained by reasoning about agents' equilibrium behavior in the game perceived by the ego agent. However, in general, no two agents need to share the same pool of players during planning. The influence of agents optimizing different games have been studied from the point of view of cost mismatch \cite{Chahine2023}. To the best of our knowledge, the question of games with player mismatching, or \emph{local games}, has not been addressed in the literature.
Second, with the nearest neighbor criterion as a baseline, we introduce several neighbor selection schemes based on Control Barrier Functions (CBF) \cite{ames2017control} \cite{Xiao2019}, cost sensitivity, and contribution to cost evolution. The aim is to understand how much improvement principled player selection methods can achieve with respect to the baseline, in order to make a case for the use of \emph{local games} as a scalable solution to multi-agent motion planning problems. The contributions of this paper are as follows:
\begin{figure}[t]
	\centering
	\includegraphics[width=0.45\textwidth]{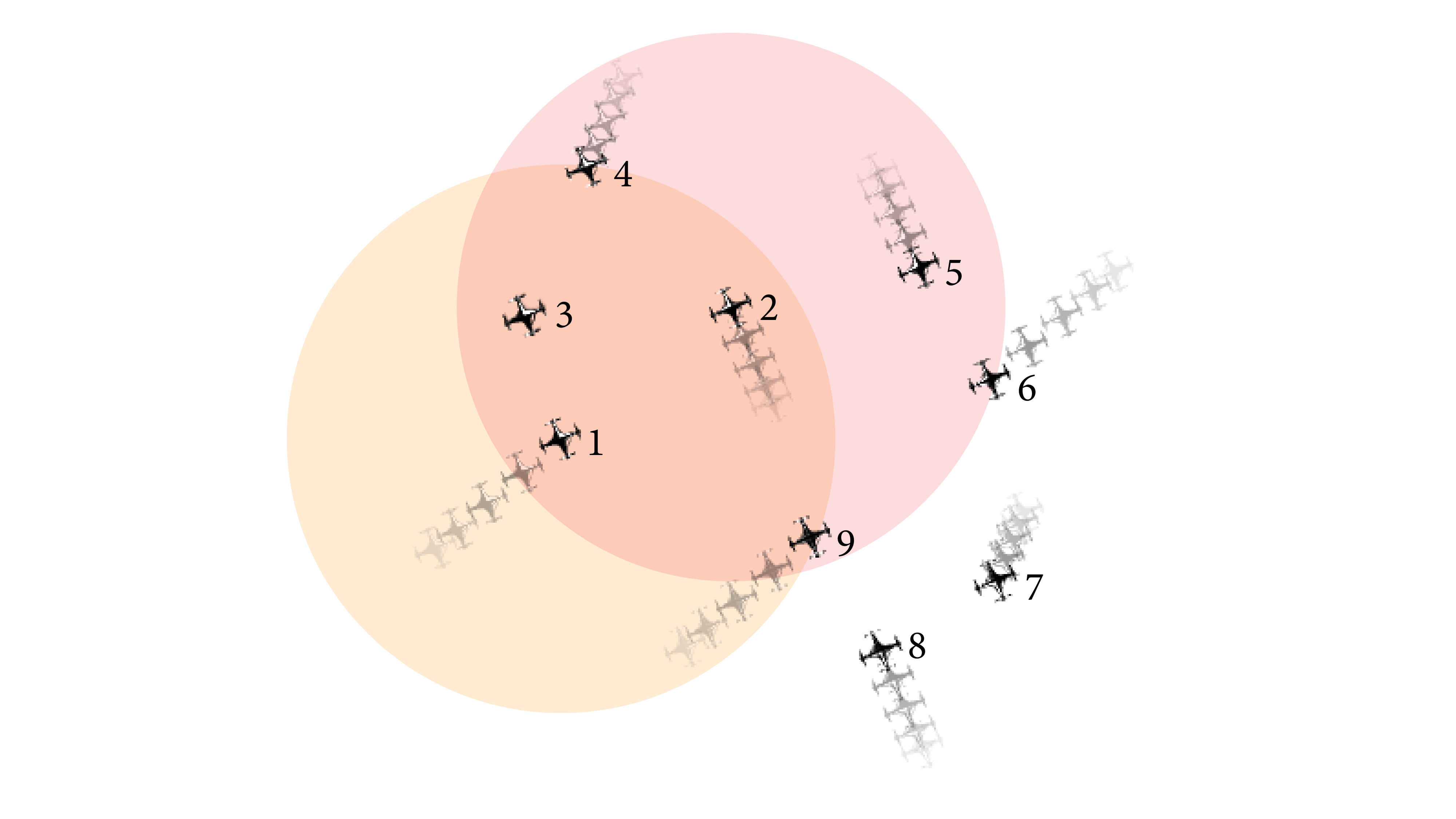} 
	\caption{In practice, robots make decisions based on local information. This can lead to agents solving games with different subsets of agents present in the scene. For example, based on a fixed radius criterion, agent 1 may consider 2, 3, 4, and 9 as neighbors in its game, while agent 2 may consider 1, 3, 4, 5, 6, and 9. How do such systems behave and can local information be leveraged to select only significantly interacting neighbors, in order to mitigate the computational burden of considering agents with little or no impact on one's actions?}
	\label{fig:teaser}
\end{figure}
\begin{itemize}
\item We investigate \emph{local games} (in which each agent considers only a small number of local neighbors) for scalable decentralized trajectory planning in congested environments.
\item We propose multiple neighbor selection schemes, amongst which is a novel use of Control Barrier Functions (CBF) for priority ranking, to overcome shortcomings of a naive nearest neighbor criterion.
\item We provide numerical analysis to study how game complexity (in number of players per game) affects anticipation capacity and safety in congested environments, and compare methods with double integrator agents in the plane and nonlinear quadrotor agents in $\mathbb{R}^3$.
\end{itemize}
The paper is organized as follows: Sec.~\ref{related_work} reviews the related work. Sec.~\ref{prob_statement} describes the problem statement. Sec.~\ref{methods} presents different proposed player section methods. Sec.~\ref{results} presents the numerical results based on simulation studies. Sec.~\ref{conclusion} makes concluding remarks and discusses the future work.

\section{Related Work} \label{related_work}

\textbf{Game-theoretic planning:}
Game-theoretic planning provides a mathematical formalism to model highly complex interactions among different agents. Interactive planning approaches can be categorized in two categories of (i) Predict-then-Plan approaches and (ii) Predict-and-Plan approaches. In Predict-then-Plan approaches, the prediction and planning are decoupled. The ego agent first predicts the other agents' behavior along a horizon in the future and then plans based on the predicted trajectories as presented in \cite{Felix2018}, and \cite{Ferranti2018}. In the second category which is predict-and-plan approaches such as game-theoretic planning, the prediction and planning steps are coupled.  The planning and prediction is performed together by finding a game theoretic equilibrium in the joint space of trajectories of all the agents in the scene as presented in \cite{spica2020real},  \cite{leCleach2022algames}, \cite{David2020}, \cite{Fisac2018}, \cite{Sadigh2016}, and \cite{Chahine2023}. With these methods, the coupled interaction or mutual influence between the other agents in the scene and the ego agent is captured. In game-theoretic planning, not only the future actions of the other agents in the scene are predicted, but the prediction is in a feedback loop with ego agent actions. The ego agent will be able to reason about how its action will affect the future reactions of other agents in the scene.

\textbf{Interaction quantification and player selection:} 
Little literature covers the question of quantifying the interactions between motion planning robots, which is important information for the selection of a subset of agents relevant to the ego robot's performance. In \cite{Schwarting2019}, the authors touch on this topic in the work's appendix. They implement an MPC game-theoretic planner limiting interactions to 4 neighboring agents in the context of autonomous driving. This limitation is justified by empirical tests but also in post analysis by computing how one agent’s utility gradient depends on another agent's inputs, i.e. a Hessian norm in the game formulation framework.
To quantify the mutual influence of interactive agents, the authors in \cite{Hyoungshick2015}, address the influential neighbor selection problem and propose a decentralized influential maximization problem by selecting k neighbors rather than arbitrary agents. The authors have introduced different types of neighbor selection schemes and have shown information on online social networks e.g. Twitter can be propagated efficiently by selecting neighbors with high propagation rate rather than those with a high number of neighbors. In \cite{Sun2016}, the authors present a reversed nearest neighbor heat map which provides influence distribution in a two-dimensional space.

In \cite{Ma2021}, the authors propose to select neighbors to conduct communication with, using deep reinforcement learning. In \cite{Niu2021}, the authors propose Graph Attention Networks to choose which agents to communicate with. In \cite{Vermeulen2017}, the authors study a range of k Nearest Neighbor (kNN) search techniques for crowd simulation. Finding kNN neighbors by querying different data structures including k-d trees, BD trees, R-trees, Voronoi diagrams, hierarchical k-mean clustering, line search and grids is studied. The query performance of these methods are compared in terms of computation time. Also, the methods are compared based on how they scale with increasing number of agents. For all approaches the number of neighboring agents is assumed to be constant $k$. In contrast, we propose to quantify the influence of other agents and take that into account to select the most influential agents as the ego agent neighbors and not a constant number of neighbors. 

% Based on vehicle-to-infrastructure (V2I) communication, the
% emergence of connected and automated vehicles \cite{Schrank2015}, \cite{Milanes2012}, \cite{Youssef2019}
% has the potential to drastically improve the performance of vehicles in traffic bottlenecks (such as merging, intersection, roundabout, etc.), ultimately
% saving energy, and avoiding congestion and accidents. Optimal control problem are used in some of these
% approaches \cite{Xiao2021}, while Model Predictive Control (MPC) techniques are employed as an alternative \cite{Ntousakis2016}, mainly to take additional constraints and disturbances into account \cite{Mohamad2018}. All the above mentioned works assume that information shared through V2I is trust-worthy. While in this work, we consider communication-based control in the presence of faulty transmissions or attacks on the communication network.  

\section{Problem Statement}\label{prob_statement}

We design and compare player selection methodologies to enable scalable receding horizon multi-agent motion planning where each robot repeatedly solves Generalized Nash Equilibrium Problems (GNEPs), updating their pool of opponents at each iteration. We thus present the GNEP framework in its general formulation. Also, a novel contribution of the work is the use of Control Barrier Functions as an interaction quantification method. The CBF framework is thus also presented in this section.

\subsection{Generalized Nash Equilibrium Problem}
A given GNEP will involve $N$ players $i \in \{1,\dots,N\}$ over a horizon of $T$ time steps. An agent $i$'s state at time step index $k$ is denoted $\mathbf{x}_k^i \in \mathbb{R}^{n^i}$ and control input $\mathbf{u}_k^i \in \mathbb{R}^{m^i}$, with dimensions of agent $i$'s state and control $n^i$ and $m^i$. Let $\mathbf{x}_k = [\mathbf{x}_k^{1, \top}, \dots, \mathbf{x}_k^{N, \top}]^\top \in \mathbb{R}^{n}$ denote the joint state and $\mathbf{u}_k = [\mathbf{u}_k^{1, \top}, \dots, \mathbf{u}_k^{N, \top}]^\top \in \mathbb{R}^{m}$ denote the joint control of all agents at time $k$, with joint dimensions $n = \sum_i n^{i}$ and $m = \sum_i m^i$. We define player $i$'s policy as $\pi^i = [\mathbf{u}_1^{1, \top}, \dots, \mathbf{u}_{T-1}^{1, \top}]^\top \in \mathbb{R}^{\Tilde{m}^i}$ where $\Tilde{m}^i = m^i (T-1)$ denotes the dimension of the entire trajectory of agent $i$'s control inputs. The notation $\neg i$ indicates all agents except $i$, for instance $\pi^{\neg i}$ represents the vector of the agents' policies except that of $i$. Also, let $X = [\mathbf{x}_2^{\top}, \dots, \mathbf{x}_N^{\top}]^\top \in \mathbb{R}^{\Tilde{n}}$, with $\Tilde{n}= n(T-1)$, denote the trajectory of joint state variables resulting from the application of the joint control inputs to the dynamical system defined by $f:\mathbb{R}^{n} \times \mathbb{R}^{m} \rightarrow \mathbb{R}^{n}$ such that,
\begin{align}
    \mathbf{x}_{k+1} = f(\mathbf{x}_{k}, \mathbf{u}_{k}). \label{dyna}
\end{align}
Over the whole trajectory we can express the above kinodynamic constraints with $\Tilde{n}$ equality constraints,
\begin{align}
    D(X,\pi^1,\dots,\pi^N) = D(X,\pi) = 0 \in \mathbb{R}^{\Tilde{n}} \label{kino}
\end{align}
The cost function of each player $i$ depends on its policy $\pi^i$ as well as on the joint state trajectory $X$, which is common to all players, such that $\forall i \in \{1,\dots,N\}$,
\begin{align}
    \mathcal{C}^i(X,\pi^i) = c_T^i(\mathbf{x}_N) + \sum_{k=1}^{T-1} c_k^i(\mathbf{x}_k, \mathbf{u}_k^{i}). \label{cost}
\end{align}
Notice that as player $i$ minimizes $\mathcal{C}^i$ with respect to $X$ and $\pi^i$, the selection of $X$ is constrained by the other
players’ strategies $\pi^{\neg i}$ and the dynamics of the joint system via (\ref{kino}). In addition, the strategy $\pi^i$ could be required to satisfy (safety) constraints that depend on the joint state trajectory $X$ as well as on the other players strategies $\pi^{\neg i}$ . This can be expressed with a set of $l$ inequality constraints,
\begin{align}
    C(X,\pi) \leq 0 \in \mathbb{R}^{l} \label{cons}
\end{align}
where $C:\mathbb{R}^{\Tilde{n}} \times \mathbb{R}^{m(T-1)}\rightarrow \mathbb{R}^l$.
The GNEP we form is the problem of minimizing (\ref{cost}) for all players $i \in \{1,\dots,N\}$ with respect to (\ref{kino}) and (\ref{cons}). More specifically,
\begin{align}
    \underset{X,\pi^i}{\min}\ & \mathcal{C}^i(X,\pi^i) \qquad \forall i \in \{1,\dots,N\} \nonumber\\
    \text{subject to }\ & D(X,\pi) = 0 \label{gnep}\\
    & C(X,\pi) \leq 0. \nonumber
\end{align}
The solution to such a dynamic game is a generalized Nash equilibrium, i.e. a policy $\hat{\pi}$ such that, $\forall i \in \{1,\dots,N\}$, $\hat{\pi}^i$ is a solution to (\ref{gnep}) with the other players' policies given by $\hat{\pi}^{\neg i}$ that is also solved by (\ref{gnep}) for all $\neg i$. As a consequence, at a Nash equilibrium point solution, no player can improve their strategy by unilaterally modifying their policy.

\subsection{Control Barrier Functions}
Control Barrier Functions (CBFs) \cite{ames2017control} are a popular control synthesis tool for safety-critical systems. CBFs can transform safety constraints of control-affine systems into state-feedback constraints that are linear in controls. With such control constraints, one can formulate convex optimization problems (CBF-based optimization) whose solutions (optimal controls) guarantee system safety.
Consider a safety constraint $h_{i,j}(X)\geq 0$ between two agents $i$ and $j$, where $h_{i,j}: \mathbb{R}^{\Tilde{n}}\rightarrow\mathbb{R}$ (recall $\Tilde{n}$ is the dimension of $X$) is continuously differentiable. Following the CBF method, we have that $h_{i,j}(X)\geq 0$ is guaranteed for all times if $h_{i,j}(X(t_0))\geq 0$ and the following CBF constraint is satisfied: 
\begin{equation} \label{eqn:bf}
    \dot h_{i,j}(X) + \alpha_1(h_{i,j}(X))\geq 0,
\end{equation}
where $\alpha_1$ is a class $\mathcal{K}$ function (strictly increasing function that passes through the origin). If the relative degree (number of times we need to take the derivative of $h_{i,j}(X)$ along dynamics until any control shows in the corresponding derivative) of $h_{i,j}(X)$ is one, then we would have control $\pi$ in $\dot h_{i,j}(X)$, i.e., we rewrite $\dot h_{i,j}(X)$ as $\dot h_{i,j}(X, \pi)$, and (\ref{eqn:bf}) can be rewritten as:
\begin{equation} \label{eqn:cbf}
    \dot h_{i,j}(X, \pi) + \kappa\alpha_1(h_{i,j}(X))\geq 0,
\end{equation}
where $\kappa > 0$ is a scalar that we added to the class $\mathcal{K}$ function. The magnitude of $\kappa$ will determine the size of the activation zone of the CBF; with empirical evidence  suggesting that it grows larger for smaller values of $\kappa$.

% When two agents get close to each other, the CBF constraint (\ref{eqn:cbf}) will usually become active in the CBF-based optimization, i.e., (\ref{eqn:cbf}) will become an equality. In other words, as two agents get closer to each other, the value of 
% \begin{equation} 
% \label{eqn:cbf_m}
% f^{CBF}_{i,j}(X,\pi,\kappa):= \dot h_{i,j}(X, \pi) + \kappa\alpha_1(h_{i,j}(X)),
% \end{equation}
% will become smaller until reaching to 0. Thus, we can take $f^{CBF}_{i,j}(X,\pi,\kappa)$ as a CBF risk metric for the collision, in which a smaller value means  a higher risk of collision. 

If the relative degree of $h_{i,j}(X)$ is $m \in\mathbb{N}$ that is larger than one, then the control terms would not appear in (\ref{eqn:cbf}).
% and in such cases (\ref{eqn:cbf_m}) is replaced by
% \begin{equation} 
% \label{eqn:bf_m}
% f^{BF}_{i,j}(X,\kappa):= \dot h_{i,j}(X) + \kappa\alpha_1(h_{i,j}(X)).
% \end{equation}
% We call it a barrier function (BF) risk metric since there is no control. Nonetheless, it can also partially capture the risk since we have additional state information (such as speeds if $h_{i,j}$ is defined over positions) compared to the nearest neighbor method.
In this case involving high-relative-degree safety constraints, high order CBF (HOCBF) methods \cite{Xiao2019} are required, by defining,
\begin{equation}
    \psi_k(X):=\dot \psi_{k-1}(X) + \alpha_k(\psi_{k-1}(X)), \quad k\in\{1,\dots,m\},
    \label{eqn:hocbf}
\end{equation}
where $\psi_0(X) = h_{i,j}(X)$, $\alpha_k, k\in\{1,\dots,m\}$ are class $\mathcal{K}$ functions, and $\psi_m(X,\pi) \equiv \psi_m(X)\geq 0$ implies $h_{i,j}(X)\geq 0$ and therefore safety. 
% Therefore, the CBF risk metric in this case is defined as:
% \begin{equation} 
% \label{eqn:hocbf_m}
% f^{CBF}_{i,j}(X,\pi):= \psi_m(X, \pi).
% \end{equation}
Again, $\kappa > 0$ can be added to each of the class $\mathcal{K}$ functions above to regulate the size of the HOCBF activation zone.

\section{Proposed player selection methods} \label{methods}

In this section, we develop the multiple ranking techniques that aim to sort agents in terms of the significance of their interactions. As the resolution of a dynamic game over some time horizon $T$, is performed on a fixed set of agents (both in number and identity), we aim to use the ranking systems to update the subset of players the ego agent takes into consideration for planning. We propose 5 methods to rank agents and to be compared along with the naive approach of considering the Euclidean nearest neighbors.

For all applications presented in this paper, we explicitly provide the structure of the cost at time step $k<T$ of a given agent $i$, appearing in \eqref{cost},
\begin{align}
    c_k^i(\mathbf{x}_k, \mathbf{u}_k^{i})
    &= (\mathbf{x}_k^i - \mathbf{x}_f^i)^\top Q_i (\mathbf{x}_k^i - \mathbf{x}_f^i) + {\mathbf{u}_k^i}^\top R_i \mathbf{u}_k^i \nonumber \\
    & \hspace{1em} + \sum_{j=1, j \neq i}^{N} \frac{\mu}{2} \max(0, R - \norm{\mathbf{x}_k^i - \mathbf{x}_k^j})^2
    \label{costex}
\end{align}
where $Q_i$ is the state quadratic cost matrix, $R_i$ the control quadratic cost matrix, $\mu$ the multiplicative constant associated with the collision avoidance cost, and $R$ the repulsion radius.
We mention that the collision avoidance term only contributes to the cost when agents are within the distance $R$. To obtain a global ranking of agents independent of the repulsion radius we can use a proxy for the collision cost by looking at the quantity,
\begin{align}
    \mathcal{C}^i_{col}(k) = \sum_{j=1, j \neq i}^{N}  \frac{\mu}{\norm{\mathbf{x}_k^i - \mathbf{x}_k^j}^2},
    \label{colcos}
\end{align}
where $\mathcal{C}^i_{col}$ represent the portion of the cost associated with collision avoidance. Indeed, in equation \eqref{costex}, the maximum operator sets the pairwise collision cost to zero as long as two agents are farther than the selected repulsion radius. Hence, two agents could be indistinguishable in terms of collision cost (i.e. both with zero contribution), while presenting very disparate collision risks with the ego agent. For instance, it is clear that an agent entering the repulsion radius at high speed in the direction of the ego agent should score much higher in terms of interaction significance compared to another that is both distant and moving away from the ego agent. The collision cost structure in \eqref{costex} is oblivious to this, but the proposed proxy \eqref{colcos}, continuous in space, can capture this difference.

\subsection{Jacobian/Hessian}

The sensitivity of an agent $i$ to another agent $j$'s actions can be defined as the derivative of the cost of agent $i$ with respect to agent $j$'s actions. In general, this metric distinguishes whether one agent has sensitivity towards another but without the other way around needing to hold. We compute it first by noticing that only the collision cost incorporates a direct short term dependence on the actions of agent $j$. 

Agent i can compute a pairwise utility Jacobian with respect to agent $j$ as follows,
\begin{align}
    J_{i,j} = \frac{\partial \mathcal{C}^i_{col}}{\partial \mathbf{u}^j}.
\end{align}

Each scalar term in the Jacobian can be approximated by the finite elements method, perturbing the controls from the previous state and forward propagating the dynamics to obtain perturbed costs.
At a given time, we can evaluate the norm of this Jacobian. This will give agent $i$ information about the sensitivity of its objective to changes in $j$'s control. Larger norms will indicate higher interaction and we thus propose a pairwise scoring function $f^J_{i,j}(X,\mu)$ such that,
\begin{align}
    f^J_{i,j}(X,\mu) = \norm{J_{i,j}(X,\mu)}.
\end{align}

Furthermore, the Hessian of that accumulated reward is the derivative of the sensitivity of agent $i$ to $j$ with respect to agent $i$'s actions. In other words, it expresses how much agent $i$'s actions will change if agent $j$'s actions change. 
Agent $i$ can compute a pairwise utility Hessian with respect to agent $j$ as follows,
\begin{align}
    H_{i,j} = \frac{\partial^2 \mathcal{C}_{col}^i}{\partial \mathbf{u}^i \partial \mathbf{u}^j}.
\end{align}
Again, we can evaluate the norm of this Hessian using finite elements. This can be an even more intricate indicator, since the cost might be sensitive to changes in $j$'s actions without $i$ being able to react to it. The Hessian norm will remain small in such a case. In contrast, larger norms will indicate higher interaction and we thus propose a pairwise scoring function $f^H_{i,j}(X,\mu)$ such that,
\begin{align}
    f^H_{i,j}(X,\mu) = \norm{H_{i,j}(X,\mu)},
\end{align}
where we use the Frobenius norm.

\subsection{Cost evolution} 

Another approach to quantify interaction can be based on the cost impact agents have on each other. Accordingly, greedy ranking can be used to limit the impact of the highest contributing agents. Statically, the highest cost is associated to the nearest neighbor. However, dynamically we can consider looking at which agent has had the biggest recent impact on cost increase. This can be achieved based on position observations only, although we implicitly take into account dynamics as we estimate the additional contribution of each agent over the past time step. Indeed, we rank agents by how fast their contribution increases the ego collision avoidance cost, i.e. by evaluating the pairwise quantity, 
\begin{align}
    f^{CE}_{i,j}(X) = \frac{\mu}{\norm{\mathbf{x}_{k}^i - \mathbf{x}_k^j}^2} - \frac{\mu}{\norm{\mathbf{x}_{k-1}^i - \mathbf{x}_{k-1}^j}^2}.
    \label{costevo}
\end{align}

Negative contributions indicate the other agent is getting farther away from the ego agent, and can more easily be ignored in planning. Another desirable feature of this metric is that for equal distance reductions, agents that are closer will contribute more to the cost increase, and will thus be higher up the interaction ranking.

\subsection{Control Barrier Function}
In this work, we use CBFs as a risk metric evaluation tool that can be used to select the most safety-threatening agents in local games. This tool is useful since CBFs usually have activation zones around the unsafe sets, and these activation zones are larger than the unsafe sets (the size depends on the CBF parameters), which means that CBFs can help to predict collisions before they may happen, as previously discussed in Sec.~\ref{prob_statement}. Between each pair of agents $i$ and $j$, given the collision radius to avoid violating, we can define a zeroth order barrier function, capturing the pairwise collision avoidance constraint, as follows,
\begin{align}
    h_{i,j}(X) &= \norm{\mathbf{x}^i - \mathbf{x}^j}^2 - R^2 %\nonumber \\
    %&= \sum_{d=1}^n (p_d^i - p_d^j)^2 - R^2
\end{align}
its first Lie derivative (along the trajectory) is with respect to time, thus,
\begin{align}
    \dot{h}_{i,j}(X) &= 2 (\mathbf{x}^i - \mathbf{x}^j)^\top (\mathbf{v}^i - \mathbf{v}^j)
    %\sum_{d=1}^n 2 (p_d^i - p_d^j)(v_d^i - v_d^j)
\end{align}
With the choice of linear class $\mathcal{K}$ functions in \eqref{eqn:cbf} and \eqref{eqn:hocbf}, a first order pairwise barrier function can thus be obtained,
\begin{align}
    f^{BF}_{i,j}(X,\kappa) = \dot{h}_{i,j}(X) + \kappa h_{i,j}(X)
\end{align}
with $\kappa>0$.
We can repeat the process to obtain a higher order barrier function in which the control input appears. For motion dynamics considered in this paper (both double intergrator and quadrotor), the order required is 2. We thus derive the second derivative,
\begin{align}
    \ddot{h}_{i,j}(X,\pi) = 2 [(\mathbf{v}^i - \mathbf{v}^j)^\top (\mathbf{v}^i - \mathbf{v}^j) + (\mathbf{x}^i - \mathbf{x}^j)^\top (\mathbf{a}^i - \mathbf{a}^j)]
    %&= 2 \sum_{d=1}^n (v_d^i - v_d^j)^2 + (p_d^i - p_d^j)(u_d^i - u_d^j)
\end{align}
where the acceleration vector of a given agent $i$, denoted above by $\mathbf{a}^i$, is a known function of its control input.
Hence, we obtain a pairwise Control Barrier Function,
\begin{align}
    f^{CBF}_{i,j}(X,\pi,\kappa) = \ddot{h}_{i,j}(X,\pi) + 2\kappa \dot{h}_{i,j}(X) + \kappa^2 h_{i,j}(X)
\end{align}
with $\kappa>0$.
Both the BF and CBF constraints requires $f^{BF}_{i,j}(X) \geq 0$  and $f^{CBF}_{i,j}(X,\pi) \geq 0$ for the $(i,j)$ pair of agents to avoid collision. Thus, larger values will correspond to pairs of agents who are far from activating the collision constraint and that are subsequently unlikely to have significant interaction. The barrier functions thus constitute a natural ranking method for planning priority.

\subsection{Methods summary}

We have thus proposed 5 methods in addition to the naive nearest neighbor approach, two of which are based on the sensitivity analysis of the cost function with respect to other agents' control (Jacobian and Hessian), another, adopting a greedy perspective consisting in dealing with agents that contribute to a cost increase at the biggest rates (Cost Evolution), and, finally, two more inspired by control theory and specifically barrier functions (BF and CBF). The computational requirements and characteristics of each method are presented in Table \ref{tab:meth_sum}.

\begin{table}[!t]
\vspace{0.2cm}
\centering
\caption{Characteristics of considered player selection methods. Ticks/crosses indicate whether or not the method requires the associated information/computation. We consider access to positions (\textbf{pos}), velocities (\textbf{vel}), control inputs and dynamics (\textbf{c\&d}), in addition to need for forward integration of dynamics in time (\textbf{f-i}) and tuning (\textbf{tun}).}
\begin{tabular}{l c c c c c}
\toprule
\textbf{} & \textbf{pos} & \textbf{vel} & \textbf{c\&d}  & \textbf{f-i} & \textbf{tun}\\ 
\midrule
\textsc{Nearest Neighbor} & \cmark & \xmark & \xmark & \xmark & \xmark\\
\textsc{Cost Evolution} & \cmark & \xmark & \xmark &\xmark & \xmark \\
\textsc{Jacobian} & \cmark & \cmark & \cmark & \cmark & \xmark\\
\textsc{Hessian} & \cmark & \cmark & \cmark & \cmark & \xmark\\
\textsc{BF} & \cmark & \cmark & \xmark & \xmark & \cmark\\
\textsc{CBF} & \cmark & \cmark & \cmark & \xmark & \cmark\\
\bottomrule
\end{tabular}
\label{tab:meth_sum}
\end{table}

\section{Simulations} \label{results}

We compare our proposed player selection criteria in three scenarios with increasing complexity. All tests consist in having the agents navigate from a start position to a random final position on a grid, minimizing their control input and simultaneously attempting to avoid other agents. This spatial architecture is interesting as agents have to deal with potential collisions from all sides, at least on the inside of the grid. The crucial computationally limiting factor is $p$, the number of players considered by the ego agent at each planning step, in other words the size of the $(p+1)$-player game considered. We aim to understand how the increase in game complexity, by incorporating additional players, impacts an agents capacity to safely navigate the environment. We assume all agents plan using the receding horizon game theoretic planner with the same $p$ and the same player selection method for each run.
The performance metric we select to differentiate the methods is the minimum distance between any pair of agents along the joint trajectory.

\subsection{Simulations setup}

The dynamic games solver we use is ALGAMES \cite{leCleach2022algames}, which is an open-source solver for game theoretic planning problems with multiple actors and general nonlinear state and input constraints.
First, we test the methods for double integrator dynamics in the plane, considering a 3x3 grid of agents. This smaller setup, though less congested, will allow us to compare the methods' performance in avoiding collisions against the full game implementation, i.e. one with all other agents considered during planning. We extend this experiment to a larger 5x5 grid, increasing the congestion and number of interactions occurring during navigation. We finally test on an even larger cubic grid of 3x3x3 agents following 12 dimensional quadrotor dynamics to exhibit the advantages of our method in real world flight applications. The tuning of $\kappa$, appearing in the barrier functions is done separately for the BF and CBF functions as well as for the chosen dynamics. However in all 4 cases we converge to a close value and thus fix $\kappa = 5.0$ in all experiments.

% \subsubsection{Constraints}
% The dynamics constraints at time $k$ are given by,
% \begin{align}
%     x_{k+1} = f(x_k, u_k) 
% \end{align}
% We also enforce collision-avoidance constraints on the trajectories, by modelling collision zones of the vehicles by circles or radius $r$, such that, at any time step $k$,
% \begin{align}
%     \norm{x^i_k - x^j_k}_2^2 \geq r^2, \quad \forall i,j \in \{1,\dots,N\}
% \end{align}
% In addition, we require the vehicles to remain on the road, by constraining the distance between each vehicle and the closest point $q$ on each boundary $b$ to remain larger than the collision radius $r$,
% \begin{align}
%     \norm{x^i_k - q_b}_2^2 \geq r^2, \quad \forall b, \forall i \in \{1,\dots,N\}
% \end{align}
% Thus, the autonomous driving problem is formalized via non-convex and non-linear coupled constraints.

% \subsubsection{Cost function}
% The cost structure considered is quadratic, penalizing the distance to the desired final state $x_f$ and the use of controls,
% \begin{align}
%     J^i(X,\pi^i) &= \sum_{k-1}^{T-1}\frac{1}{2}(x_k-x_f)^\top Q (x_k-x_f) + \frac{1}{2} u_k^{i,\top} R u_k^i \nonumber\\ & \hspace{1em} + \frac{1}{2}(x_T-x_f)^\top Q_f (x_t-x_f)
% \end{align}
% This cost function depends only on the decision variables of vehicle $i$. Players’ behaviors are thus only coupled through the collision constraints.

\subsection{Double integrator}

We first consider agents evolving in the 2D plane according to double integrator dynamics. The state of a vehicle comprises of its 2D positions and velocities and the control input comprises of the 2D accelerations.

\subsubsection{3x3 grid}

\begin{figure}[!tb] \center \vspace{0.2cm}
\includegraphics[width=0.45\textwidth]{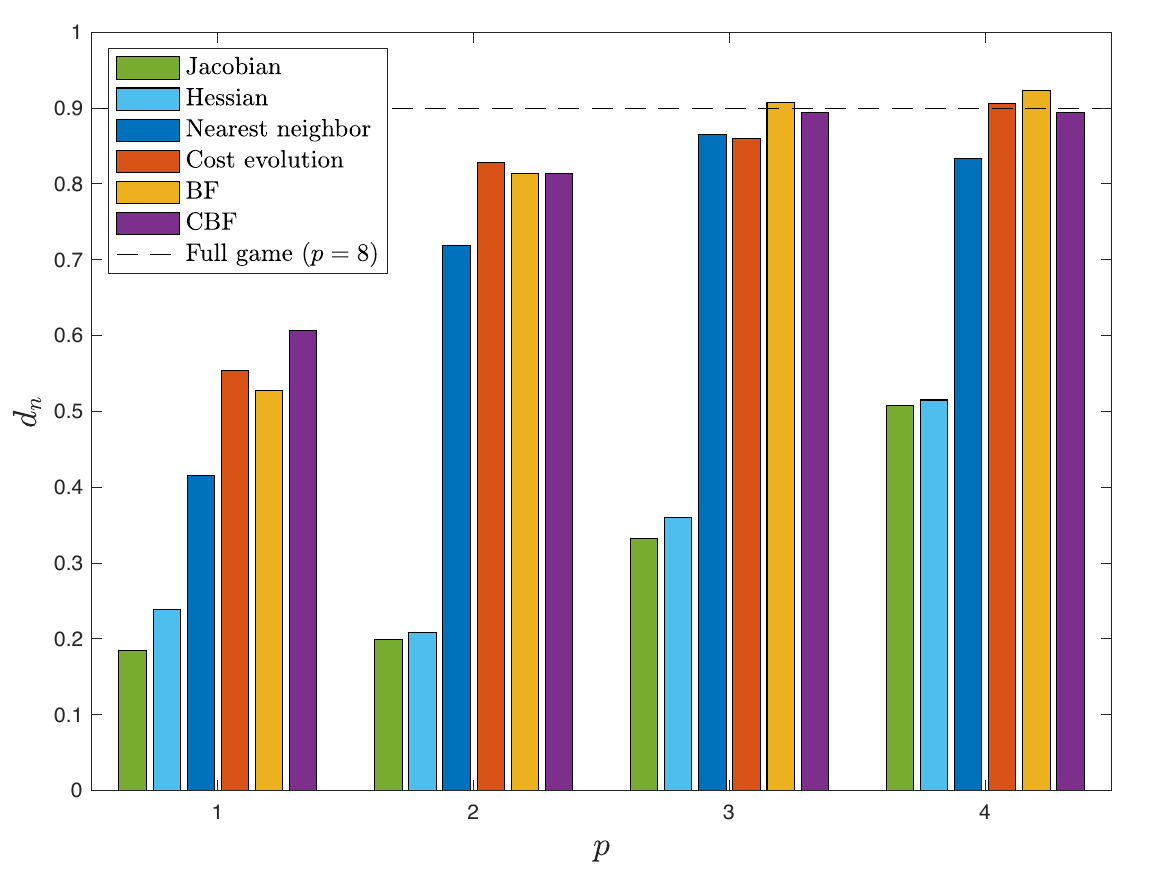}
	\caption{3x3 grid: Mean normalized minimum distances between any pair of agents with $p$ other participants using the Jacobian, Hessian, Nearest neighbor, Cost evolution, BF, and CBF schemes. The dotted black line is the value obtained by solving the full 9-player game (also averaged over runs). The normalization factor is the constraint activation radius defined in \eqref{costex}.} \vspace{-0.2cm}
	\label{fie:di9_res}
\end{figure}

The setup is small enough to permit the complete resolution of the game in reasonable time offline, i.e. taking all other 8 agents into account when planning at each receding horizon time step. Thus, we can obtain a ground truth baseline against which to compare the performance of local games with different player selection criteria. We repeat the experiment 20 times, randomizing the desired final positions of agents on the grid, for $p = 1,\dots,4$, and for all 6 player selection methods, as well as solving the entire 9-player game. The minimum pairwise distances are averaged over the runs, the results are presented in Fig. \ref{fie:di9_res}.

First, we notice that both methods (Jacobian and Hessian) based on the partial differentiation of the ego agent's cost with respect to controls perform even worse than the nearest neighbor approach. This could be due to the absence of dependence on the agents' trajectories and their evolution in time, taking only into account a short term relationship to a change in controls. We will abandon the further analysis of these methods in the subsequent experiments.
Also, we notice that both barrier function methods and the cost evolution criterion perform similarly well and improve performance against the nearest neighbor method, especially for games restrained to 1 or 2 opponents.
Elsewhere, the limited congestion due to the small number of agents in the scene ensures that collision avoidance performance levels comparable to those for the full game can be achieved by considering 3 or more participants.

\begin{figure}[!tb] \center \vspace{0.2cm}
\includegraphics[width=0.45\textwidth]{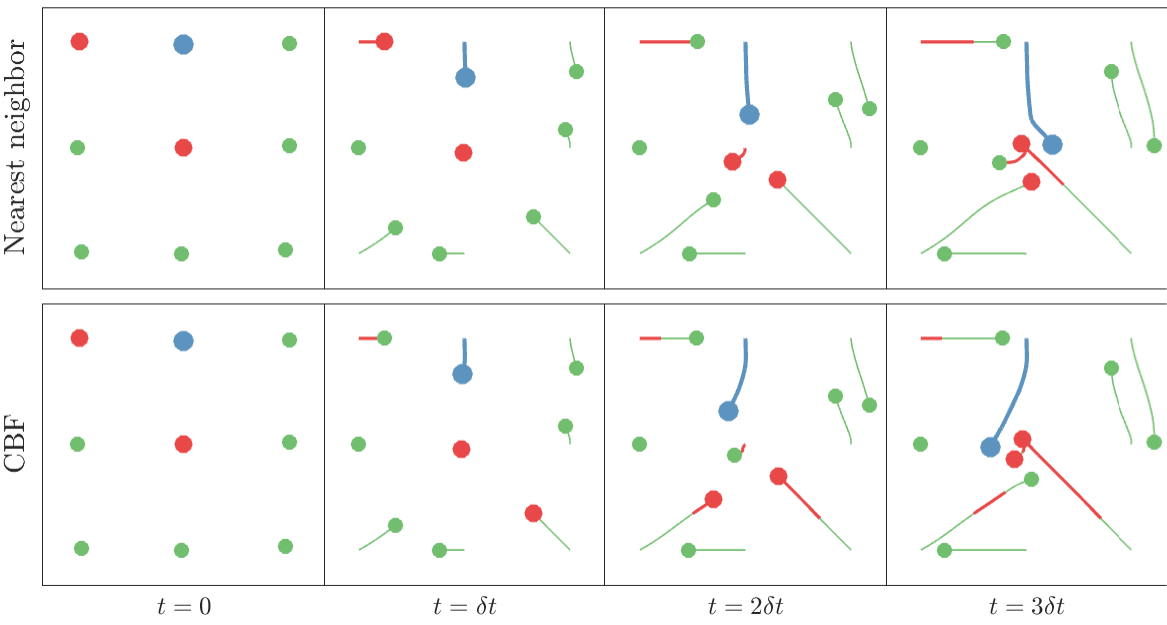}
	\caption{Nearest neighbor vs CBF player selection for $p=2$: Trajectory screenshots at 4 equally spaced time steps, the solid markers depicting instantaneous positions of the agents in the plane and the lines their past trajectories. The ego agent is in blue, agents used for planning are in red, and agents ignored in green.} \vspace{-0.0cm}
	\label{fie:nnvcbf}
\end{figure}

We can also analyse how non-static methods anticipate the future collision threat of agents in the scene and the effect of such anticipation on the safety of planned trajectories. Indeed, as depicted in Fig. \ref{fie:nnvcbf}, we can see that the myopic nearest neighbor approach can easily lead to more dangerous situations.
At time $t=\delta t$, the ego agent switches attention from the agent on the top left (benignly passing behind) to the one coming from the bottom right with whom it is on a collision course. By the time the bottom right agent is considered using the nearest neighbor approach at $t=2\delta t$, a maneuvre is required to avoid it. In the meantime, using the CBF ranking system, the bottom left agent is already included in planning as the ego agent notices they can neglect the centre agent which is near immobile. At $t=3\delta t$, the ego is in the clear using CBF, whereas it has to deal with another short notice collision avoidance maneuvre with the bottom left agent with the nearest neighbor selection method.
\subsubsection{5x5 grid}

\begin{figure}[!tb] \center \vspace{0.2cm}
\includegraphics[width=0.45\textwidth]{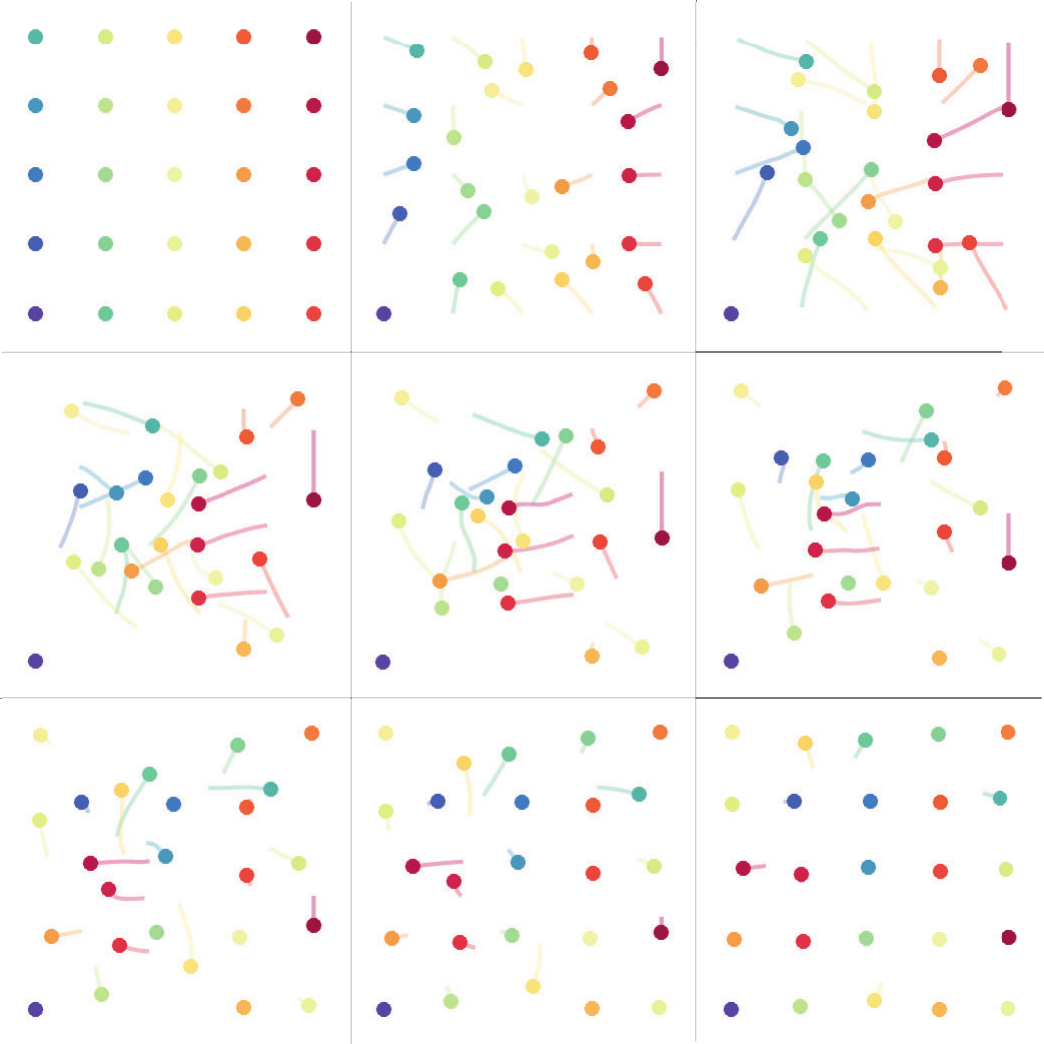}
	\caption{Sample run depicting the congested environment agents have to negotiate their way through to their target positions. Solid circles depict agents' instantaneous position, lines represent the agents trace of positions over the last 10 time steps. Time increases from left to right, top to bottom.} \vspace{-0.0cm}
	\label{fie:traj_25}
\end{figure}

\begin{figure}[!tb] \center \vspace{0.1cm}
\includegraphics[width=0.45\textwidth]{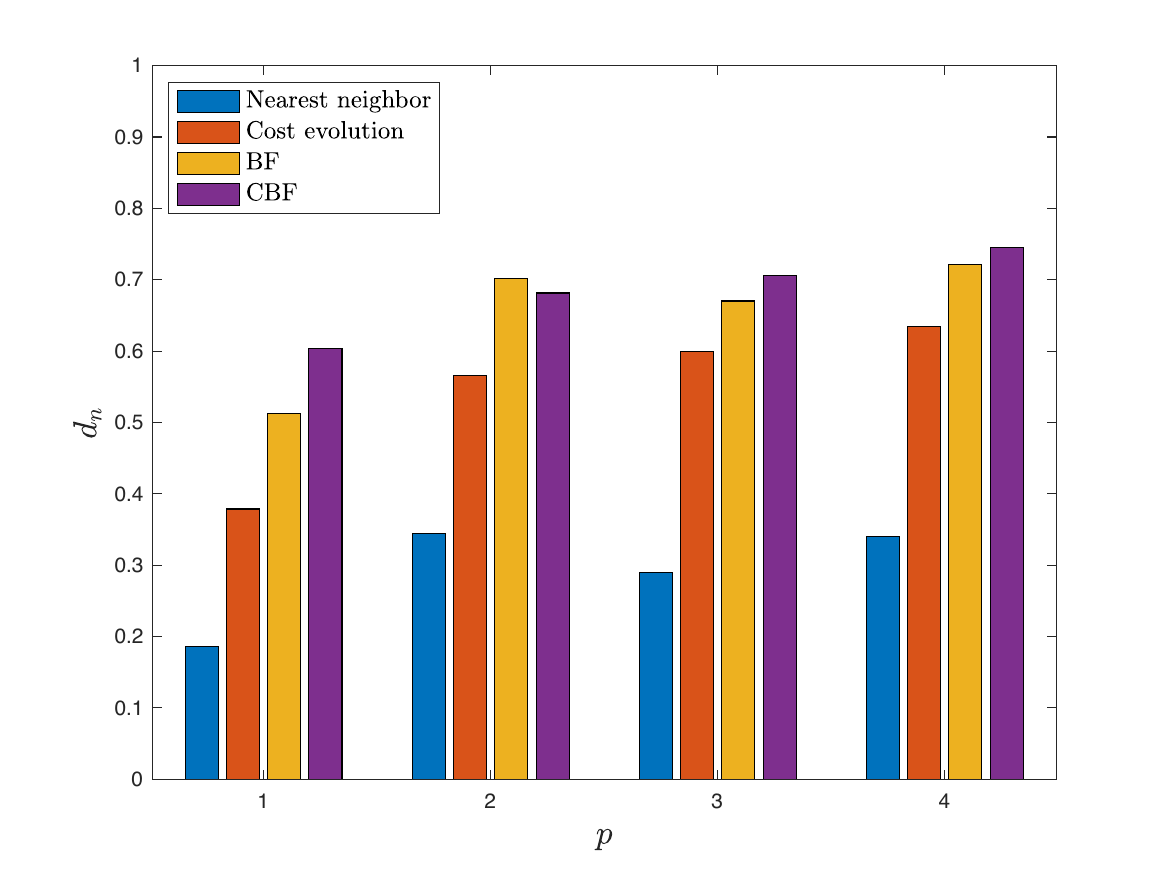}
	\caption{5x5 grid: Mean normalized minimum distances between any pair of agents with $p$ other participants using the Nearest neighbor, Cost evolution, BF, and CBF schemes. The normalization factor is the constraint activation radius defined in \eqref{costex}.} \vspace{-0.2cm}
	\label{fie:di25_res}
\end{figure}

The more crowded setup provides a platform to emphasize the differences between player selection methods. Indeed, the sample trajectory depicted in Fig. \ref{fie:traj_25} shows the large number of multi-directional robots an agent finds themselves surrounded by while navigating towards their desired positions.
Again, we repeat the experiment 20 times, randomizing the desired final positions of agents on the grid, for $p = 1,\dots,4$, and for the nearest neighbor, cost evolution, BF, and CBF methods. The minimum pairwise distances are averaged over the runs, the results are presented in Fig. \ref{fie:di25_res}.
It is clear that the increased number of agents makes this experiment more challenging than the previous. Indeed, we notice that performance levels are, as expected, globally lower. More interestingly, this experiments exhibits the limits of the myopic nearest neighbor approach, which performs poorly in avoiding other agents, even as the number of participants considered grows. Moreover, a hierarchy between the remaining methods manifests itself in crowded maneuvering. Ranking agents based on their contribution in increasing the value of an ego agent's cost function over the last time step performs consistently worse than the barrier function based methods. Indeed, this disparity is clearest for pairwise game planning ($p=1$), where the repulsion radius is violated on average by over 60\% with the cost evolution method, while the CBF criterion limits violation to under 40\%. We mention that the effect of the difference in cost is even more exacerbated as the dependency is quadratic as per \eqref{costex}. Although this gap decreases as $p$ grows, it remains significant enough to consider the cost evolution criterion inferior to the barrier function methods. In fact, the latter two offer similar performance for $p \geq 2$, with the added information of CBF seamingly giving it a slight advantage. A nice advantage of the CBF metric, is that it seems to suffer less from fluctuation in performance with respect to the number of agents considered than the simpler BF, and presents consistent improvement as $p$ grows. 

\subsection{Quadrotor dynamics}

Next, we retain the Cost Evolution and CBF methods and test them against the Nearest Neighbor baseline with agents evolving according to quadrotor dynamics randomly exchanging positions on a 3x3x3 grid. The state of an agent comprises of the Euclidean positions $\mathbf{x} = [x,y,z]^\top$, Euler angles $\mathbf{q} = [\phi,\theta,\psi]^\top$, translation velocities $\mathbf{v} = [v_x,v_y,v_z]^\top$, and angular rates $\mathbf{\omega} = [\omega_\phi,\omega_\theta,\omega_\psi]^\top$ constituting a 12 dimensional vector $[\mathbf{x}^\top, \mathbf{q}^\top,\mathbf{v}^\top,\mathbf{\omega}^\top]^\top$. The control vector is the power input to each of the 4 motors $\mathbf{u} = [w_1,w_2,w_3,w_4]$. The dynamics are given by,
\begin{align}
    \mathbf{\dot{x}} &= \mathbf{v}, \quad
    &&\mathbf{\dot{q}} = R(\phi,\theta,\psi)\mathbf{\omega}, \nonumber \\
    \mathbf{\dot{v}} &= \frac{\mathbf{F}}{m}, \quad
    &&\mathbf{\dot{\omega}} = I^{-1}(\mathbf{\tau} - \mathbf{\omega} \times (I\mathbf{\omega})),
\end{align}
with $R(\phi,\theta,\psi)$ the rotation matrix between the drone body and world frames, $m$ the mass, and $I$ the inertia of the drone \cite{Hoffmann2007}. The force $\mathbf{F}$ and moment $\mathbf{\tau}$ vectors are given by,
\begin{align*}
    \mathbf{F} &= m\mathbf{g} + R(\phi,\theta,\psi)[0,0,k_f (w_1+w_2+w_3+w_4)], \\
    \mathbf{\tau} &= [L k_f (w_2-w_4), L k_f (w_3-w_1), k_m (w_1-w_2+w_3-w_4)]
\end{align*}
with $\mathbf{g} = [0,0,-g]$ the gravity vector, $k_f$ and $k_m$ the motor force and torque constants, and $L$ the distance between the motors.

\begin{figure}[!tb] \center \vspace{0.2cm} 
\includegraphics[width=0.45\textwidth]{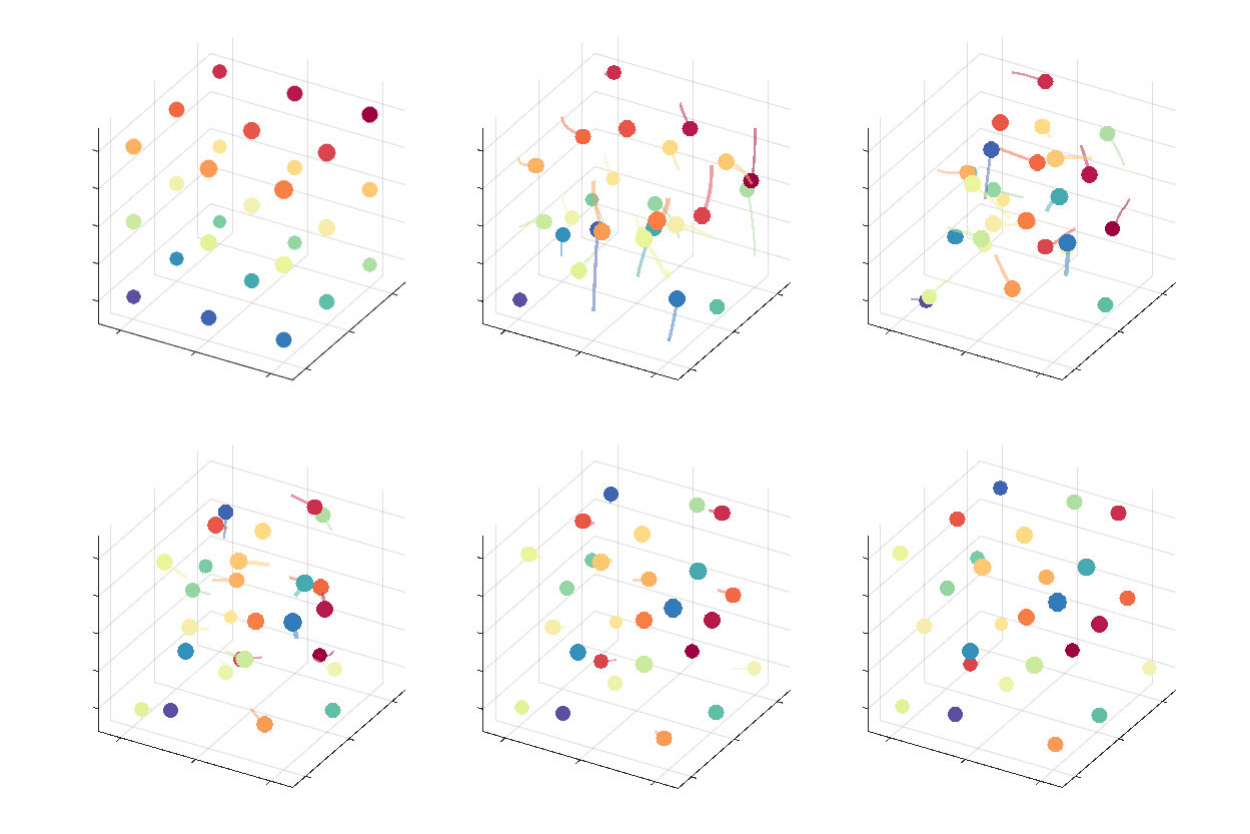}
	\caption{Sample run depicting 3x3x3 grid swap with quadrotor dynamics. Solid circles depict agents' instantaneous position, lines represent the agents trace of positions over the last 10 time steps. Time increases from left to right, top to bottom.} \vspace{-0.0cm}
	\label{fie:traj_quad}
\end{figure}

\begin{figure}[!tb] \center \vspace{0.2cm}
\includegraphics[width=0.45\textwidth]{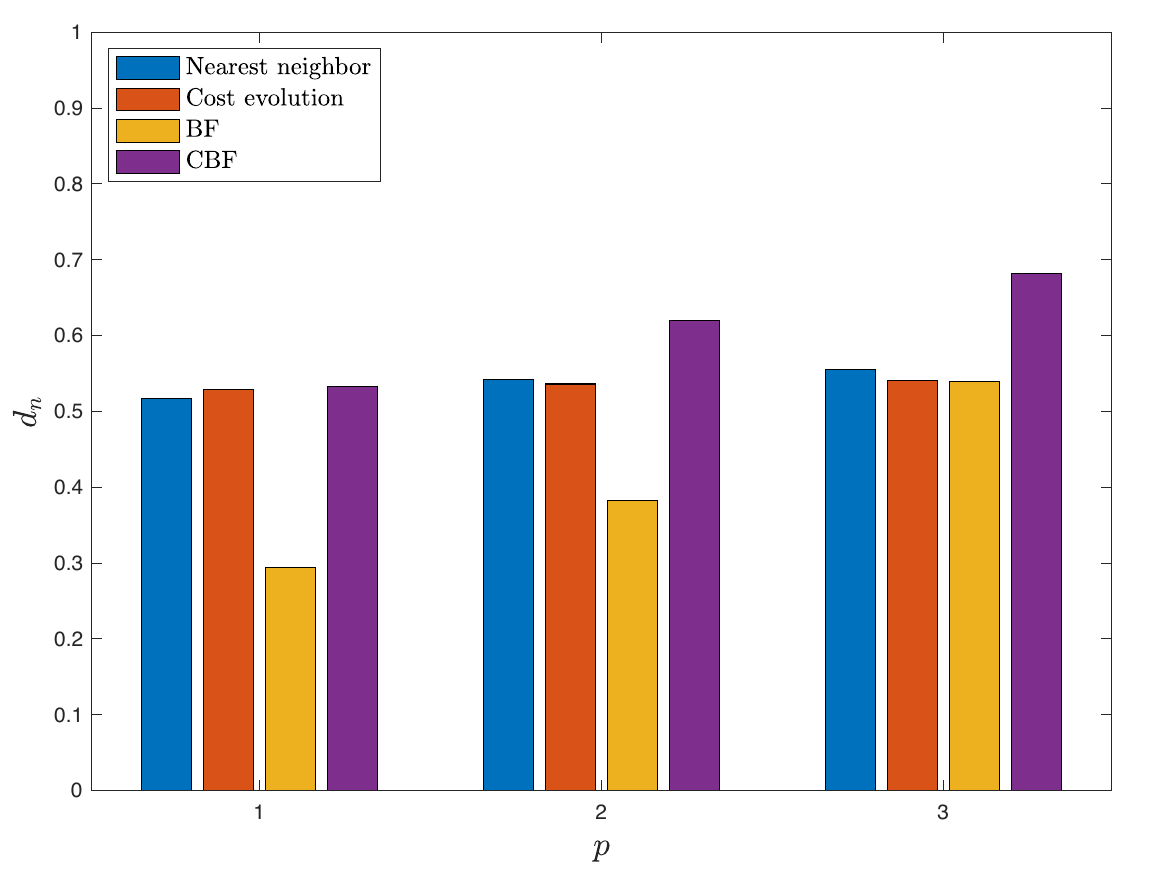}
	\caption{Mean normalized minimum distances between any pair of quadrotors versus $p$. Nearest neighbor, Cost evolution and CBF schemes. The normalization factor is the constraint activation radius defined in \eqref{costex}.} \vspace{-0.2cm}
	\label{fie:quad_res}
\end{figure}

We run 20 batches of simulations, each with a different random final state configuration over the 3x3x3 grid. For each, we compare the Nearest Neighbor approach with the player selection schemes based on the Cost Evolution Criterion and the CBF metric. Each experiment is repeated for games of fixed sizes 2 to 4. For every configuration, the minimum pairwise distances over all trajectories are computed, averaged over random final positions on the grid and presented in Figure \ref{fie:quad_res}.
Surprisingly, the nearest neighbor and cost evolution schemes perform consistently as the number of agents per game increases. Also, both methods seem to do better than the BF selection criterion. The latter, however, behaves as expected; improving performance as the game size grows, although its performance is substantially worse on smaller games. The CBF method also consistently improves collision avoidance as the games grow in size. Again, it outperforms the other methods, and does so significantly for $p \geq 2$.

\section{Conclusion}
\label{conclusion}
We explored the concept of local game-theoretic motion planning that scales with the number of agents in the scene. To maintain satisfactory collision avoidance performance while ensuring scalability, we designed local planners selecting agents with the strongest interaction as neighbors. To this end we proposed different player selection schemes and quantified the interactions of the ego agent with others in the scene. We compared various methods including Jacobian, Hessian, Nearest Neighbor, Cost Evolution, Barrier Function, and Control Barrier Function. Based on the numerical results, Control Barrier Functions consistently outperform the rest of the player selection methods, and prove to be the strongest candidate for principled player selection for local games.
% In the future, we want to investigate how estimation based on smaller subsets of players can result in faster predictions (covering all the agents with multiple smaller subsets). Also, we want to investigate parallel estimation using these smaller groups, for example a groups of two and investigate how to combine the estimates. 

\bibliography{biblo.bib}
\addtolength{\textheight}{-12cm}   % This command serves to balance the column lengths

%%%%%%%%%%%%%%%%%%%%%%%%%%%%%%%%%%%%%%%%%%%%%%%%%%%%%%%%%%%%%%%%%%%%%%%%%%%%%%%%

%%%%%%%%%%%%%%%%%%%%%%%%%%%%%%%%%%%%%%%%%%%%%%%%%%%%%%%%%%%%%%%%%%%%%%%%%%%%%%%%

%%%%%%%%%%%%%%%%%%%%%%%%%%%%%%%%%%%%%%%%%%%%%%%%%%%%%%%%%%%%%%%%%%%%%%%%%%%%%%%%

\end{document}